\documentclass[10pt,twocolumn,letterpaper]{article}

\usepackage{cvpr}
\usepackage{times}
\usepackage{epsfig}
\usepackage{graphicx}
\usepackage{amsmath}
\usepackage{amssymb}
\usepackage{comment}
\usepackage{multirow}
\usepackage{xcolor}
\usepackage{pifont}
\usepackage{colortbl}  
\usepackage{booktabs}  

\newcommand{\proposed}{CityFlow\xspace}

\usepackage[pagebackref=true,breaklinks=true,letterpaper=true,colorlinks,bookmarks=false]{hyperref}

\cvprfinalcopy 

\definecolor{lightgray}{RGB}{220,220,220}
\definecolor{darkblue}{RGB}{0,0,127}
\definecolor{darkgreen}{RGB}{0,127,0}
\definecolor{darkred}{RGB}{200,0,0}
\def\greencheckmark{\textcolor{darkgreen}{\checkmark}}
\def\redxmark{\textcolor{darkred}{\ding{55}}}  
\def\cvprPaperID{6334} 

\begin{document}

\title{\proposed: A City-Scale Benchmark for Multi-Target \\Multi-Camera Vehicle Tracking and Re-Identification}

\author{Zheng Tang\textsuperscript{1\thanks{Work done during an internship at NVIDIA.} } \quad Milind Naphade\textsuperscript{2} \quad Ming-Yu Liu\textsuperscript{2} \quad Xiaodong Yang\textsuperscript{2} \quad Stan Birchfield\textsuperscript{2}\\
Shuo Wang\textsuperscript{2} \quad Ratnesh Kumar\textsuperscript{2} \quad David Anastasiu\textsuperscript{3} \quad Jenq-Neng Hwang\textsuperscript{1}\\
\textsuperscript{1}University of Washington \quad \textsuperscript{2}NVIDIA \quad \textsuperscript{3}San Jose State University
}

\maketitle

\begin{abstract}

Urban traffic optimization using traffic cameras as sensors is driving the need to advance state-of-the-art multi-target multi-camera (MTMC) tracking. 
This work introduces \proposed , a city-scale traffic camera dataset consisting of more than 3 hours of synchronized HD videos from 40 cameras across 10 intersections, with the longest distance between two simultaneous cameras being 2.5~km. 
To the best of our knowledge, \proposed\ is the largest-scale dataset in terms of spatial coverage and the number of cameras/videos in an urban environment. 
The dataset contains more than 200K annotated bounding boxes covering a wide range of scenes, viewing angles, vehicle models, and urban traffic flow conditions. 
Camera geometry and calibration information are provided to aid spatio-temporal analysis. 
In addition, a subset of the benchmark is made available for the task of image-based vehicle re-identification (ReID). 
We conducted an extensive experimental evaluation of baselines/state-of-the-art approaches in MTMC tracking, multi-target single-camera (MTSC) tracking, object detection, and image-based ReID on this dataset, analyzing the impact of different network architectures, loss functions, spatio-temporal models and their combinations on task effectiveness. 
An evaluation server is launched with the release of our benchmark at the \href{https://www.aicitychallenge.org/}{2019 AI City Challenge} that allows researchers to compare the performance of their newest techniques. 
We expect this dataset to catalyze research in this field, propel the state-of-the-art forward, and lead to deployed traffic optimization(s) in the real world. 

\end{abstract}

\begin{figure}[t]
\begin{center}
\includegraphics[width=1.0\linewidth]{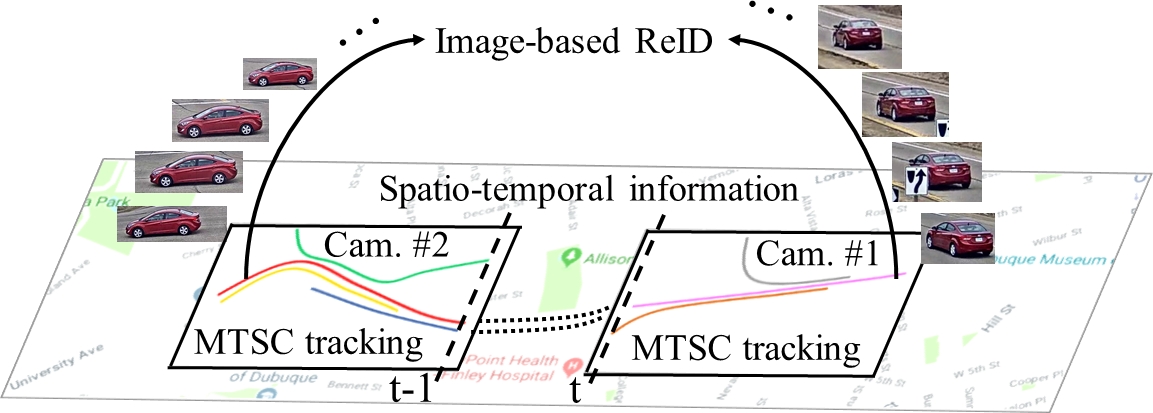}
\end{center}
   \caption{MTMC tracking combines MTSC tracking, image-based ReID, and spatio-temporal information. The colored curves in Camera \#1 and Camera \#2 are trajectories from MTSC tracking to be linked across cameras by visual-spatio-temporal association.}
\label{fig:illustration}
\end{figure}

\section{Introduction}


The opportunity for cities to use traffic cameras as citywide sensors in optimizing flows and managing disruptions is immense. 
Where we are lacking is our ability to track vehicles over large areas that span multiple cameras at different intersections in all weather conditions. 
To achieve this goal, one has to address three distinct but closely related research problems:  
1) Detection and tracking of targets within a single camera, known as multi-target single-camera (MTSC) tracking;
2) Re-identification of targets across multiple cameras, known as ReID; and
3) Detection and tracking of targets across a network of cameras, known as multi-target multi-camera (MTMC) tracking. MTMC tracking can be regarded as the combination of MTSC tracking within cameras and image-based ReID with spatio-temporal information to connect target trajectories between cameras, as illustrated in Fig.~\ref{fig:illustration}.

Much attention has been paid in recent years to the problem of \textit{person-based} ReID and MTMC tracking \cite{Zheng15a, Ristani16, Zheng17a, Zheng18, Li14, Wei12, Gray08, Hirzer11, Cheng11, Zheng16, Ristani16, Wu18, Chen15, Zheng19}.  
There have also been some works on providing datasets for \textit{vehicle-based} ReID
\cite{Liu16a, Liu16b, Yan17}. 
Although the state-of-the-art performance on these latter datasets has been improved by recent approaches, accuracy in this task still falls short compared to that in person ReID. 
The two main challenges in vehicle ReID are small inter-class variability and large intra-class variability, \textit{i.e.}, the variety of shapes from different viewing angles is often greater than the similarity of car models produced by various manufacturers~\cite{Em17}.
We note that, in order to preserve the privacy of drivers, captured license plate information---which otherwise would be extremely useful for vehicle ReID---should not be used~\cite{DPPA94}. 


A major limitation of existing benchmarks for object ReID (whether for people or vehicles) is the limited spatial coverage and small number of cameras used---this is a disconnect from the city-scale deployment level they need to operate at. 
In the two person-based benchmarks that have camera geometry available, DukeMTMC~\cite{Ristani16, Wu18} and NLPR\_MCT~\cite{Chen15}, the cameras span less than $300\times300$~m$^2$, with only 6 and 8 views, respectively. 
The vehicle-based ReID benchmarks, such as VeRi-776~\cite{Liu16a}, VehicleID~\cite{Liu16b}, and PKU-VD~\cite{Yan17}, do not provide the original videos or camera calibration information.
Rather, such datasets assume that MTSC tracking is perfect, \textit{i.e.}, image signatures are grouped by correct identities within each camera, which is not reflective of real tracking systems.
Moreover, in the latter datasets~\cite{Liu16b, Yan17}, only the front and back views of the vehicles are available, thus limiting the variability due to viewpoint.   
None of these existing benchmarks for vehicle ReID facilitate research in MTMC vehicle tracking. 

\begin{figure}[t]
\begin{center}
\includegraphics[width=0.9\linewidth]{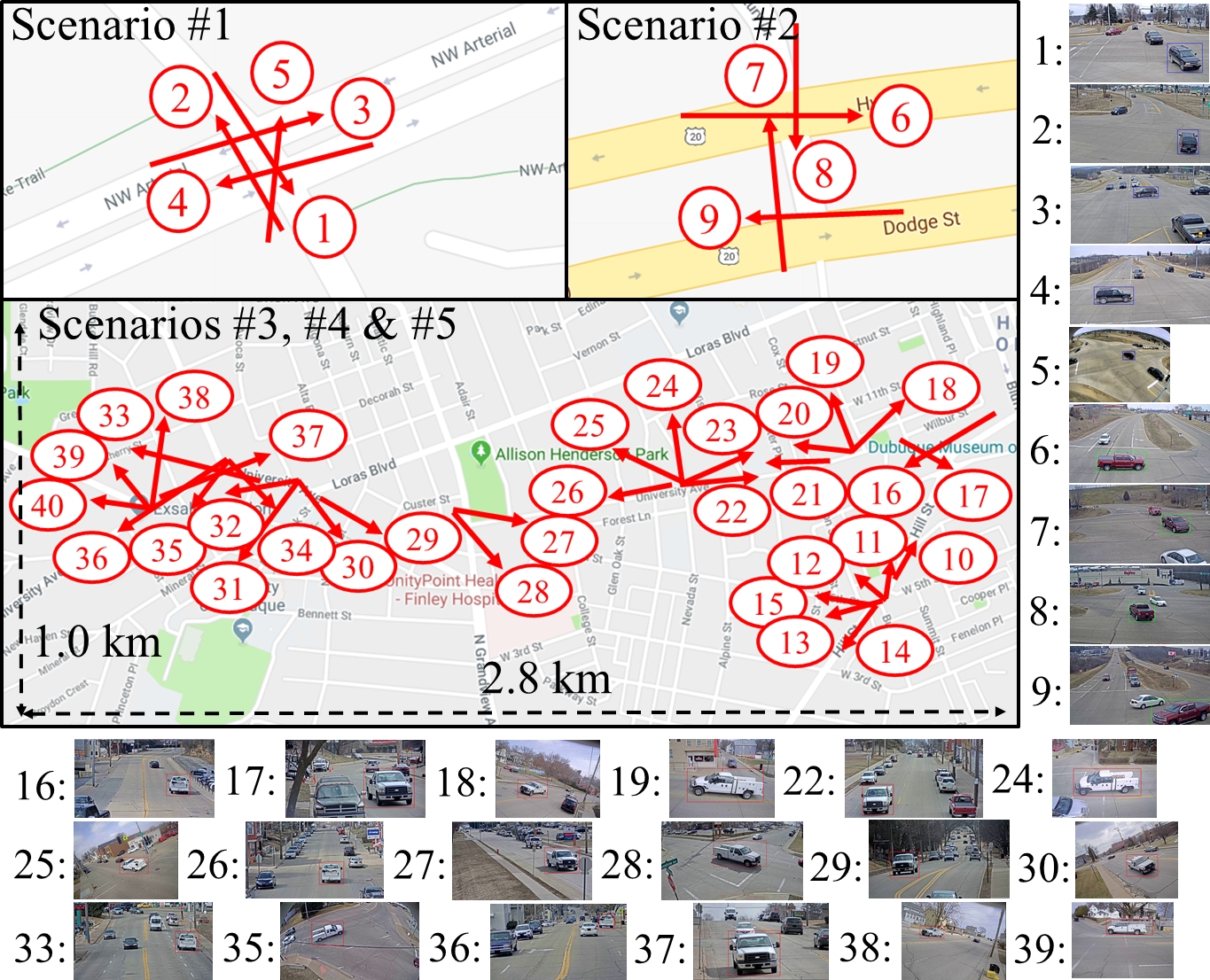}
\end{center}
   \caption{The urban environment and camera distribution of the proposed dataset. The red arrows denote the locations and directions of cameras. Some examples of camera views are shown. Note that, different from other vehicle ReID benchmarks, the original videos and calibration information will be available.}
\label{fig:map}
\end{figure}

In this paper, we present a new benchmark---called \proposed ---for city-scale MTMC vehicle tracking, which is described in Fig.~\ref{fig:map}. 
To our knowledge, this is the first benchmark at city scale for MTMC tracking in terms of the number of cameras, the nature of the synchronized high-quality videos, and the large spatial expanse captured by the dataset. 
In contrast to the previous benchmarks, \proposed\ contains the largest number of cameras (40) from a large number of intersections (10) in a mid-sized U.S.\ city, and covering a variety of scenes such as city streets, residential areas, and highways. 
Traffic videos at intersections present complex challenges as well as significant opportunities for video analysis, going beyond traffic flow optimization to pedestrian safety. 
Over 200K bounding boxes were carefully labeled, and the homography matrices that relate pixel locations to GPS coordinates are available to enable precise spatial localization. 
Similar to the person-based MTMC tracking benchmarks~\cite{Zheng16, Ristani16, Wu18}, we also provide a subset of the dataset for image-based vehicle ReID.  
In this paper, we describe our benchmark along with extensive experiments with many baselines/state-of-the-art approaches in image-based ReID, object detection, MTSC tracking, and MTMC tracking. 
To further advance the state-of-the-art in both ReID and MTMC tracking, an evaluation server is also released to the research community. 

\begin{table*}
\begin{center}
\begin{tabular}{lcccccccc}
\toprule
& & Benchmark & \# cameras & \# boxes & \# boxes/ID & Video & Geom. & Multiview \\
\midrule
\multirow{11}{*}{\rotatebox{90}{person}} &
\multirow{8}{*}{\rotatebox{0}{ReID}} & 
Market1501~\cite{Zheng15a} & 6 & 32,668 & 30.8 & \redxmark & \redxmark & \greencheckmark\\
& & DukeMTMC-reID~\cite{Ristani16, Zheng17a} & 8 & 36,411 & 20.1 & \redxmark & \redxmark & \greencheckmark\\
& & MSMT17~\cite{Wei18a} & 15 & 126,441 & 21.8 & \redxmark & \redxmark & \greencheckmark\\
& & CUHK03~\cite{Li14} & 2 & 13,164 & 19.3 & \redxmark & \redxmark & \redxmark \\
& & CUHK01~\cite{Wei12} & 2 & 3,884 & 4.0 & \redxmark & \redxmark & \redxmark \\
& & VIPeR~\cite{Gray08} & 2 & 1,264 & 2.0 & \redxmark & \redxmark & \redxmark \\
& & PRID~\cite{Hirzer11} & 2 & 1,134 & 1.2 & \redxmark & \redxmark & \redxmark \\
& & CAVIAR~\cite{Cheng11} & 2 & 610 & 8.5 & \redxmark & \redxmark & \redxmark \\
\cmidrule{2-9}
& \multirow{4}{*}{\raisebox{0.8em}{\rotatebox{0}{MTMC}}} 
& MARS~\cite{Zheng16} & 6 & 1,191,003 & 944.5 & \redxmark & \redxmark & \greencheckmark\\
& & DukeMTMC~\cite{Ristani16, Wu18} & 8 & 4,077,132 & 571.2 &  \greencheckmark & \greencheckmark & \greencheckmark\\
& & NLPR\_MCT~\cite{Chen15} & 12 & 36,411 & 65.8 & \greencheckmark & \greencheckmark & \greencheckmark \\
\midrule[\heavyrulewidth]
\multirow{5}{*}{\rotatebox{90}{vehicle}} &
\multirow{4}{*}{\rotatebox{0}{ReID}} 
& VeRi-776~\cite{Liu16a} & 20 & 49,357 & 63.6 & \redxmark & \greencheckmark & \greencheckmark \\
& & VehicleID~\cite{Liu16b} & 2 & 221,763 & 8.4 & \redxmark & \redxmark & \redxmark \\
& & PKU-VD1~\cite{Yan17} & - & 846,358 & 6.0 & \redxmark & \redxmark & \redxmark \\
& & PKU-VD2~\cite{Yan17} & - & 807,260 & 10.1 & \redxmark & \redxmark & \redxmark \\
\cmidrule{2-9}
& \multirow{1}{*}{\rotatebox{0}{MTMC}} 
& \textbf{\proposed (proposed)} & \textbf{40} & \textbf{229,680} & \textbf{344.9} & \greencheckmark & \greencheckmark & \greencheckmark \\
\bottomrule
\end{tabular}
\end{center}
\caption{Publicly available benchmarks for person/vehicle image-signature-based re-identification (ReID) and video-based tracking across cameras (MTMC). For each benchmark, the table shows the number of cameras, annotated bounding boxes, and average bounding boxes per identity, as well as the availability of original videos, camera geometry, and multiple viewing angles.}
\label{tab:public}
\end{table*}

\section{Related benchmarks}

The popular publicly available benchmarks for the evaluation of person and vehicle ReID are summarized in Tab.~\ref{tab:public}. 
This table is split into blocks of image-based person ReID, video-based MTMC human tracking, image-based vehicle ReID, and video-based MTMC vehicle tracking. 

The most popular benchmarks to date for image-based person ReID are Market1501~\cite{Zheng15a}, CUHK03~\cite{Li14} and DukeMTMC-reID~\cite{Ristani16, Zheng17a}. 
Small-scale benchmarks, such as CUHK01~\cite{Wei12}, VIPeR~\cite{Gray08}, PRID~\cite{Hirzer11} and CAVIAR~\cite{Cheng11}, provide test sets only for evaluation. 
Recently, Zheng \etal released a benchmark with the largest scale to date, MSMT17~\cite{Zheng17a}. 
Most state-of-the-art approaches on these benchmarks exploit metric learning to classify object identities, where common loss functions include hard triplet loss~\cite{Hermans17}, cross entropy loss~\cite{Szegedy16}, center loss~\cite{Wen16}, \textit{etc.} 
However, due to the relatively small number of cameras in these scenarios, the domain gaps between datasets cannot be neglected, so transfer learning for domain adaptation has attracted increasing attention~\cite{Wei18a}.

On the other hand, the computation of deep learning features is costly, and thus spatio-temporal reasoning using video-level information is key to applications in the real world. 
The datasets Market1501~\cite{Zheng15a} and DukeMTMC-reID~\cite{Ristani16, Zheng17a} both have counterparts in video-based ReID, which are MARS~\cite{Zheng16} and DukeMTMC~\cite{Ristani16, Wu18}, respectively. 
Though the trajectory information is available in MARS~\cite{Zheng16}, the original videos and camera geometry are unknown to the public, and thus the trajectories cannot be associated using spatio-temporal knowledge. 
Both DukeMTMC~\cite{Ristani16, Wu18} and NLPR\_MCT~\cite{Chen15}, however, provide camera network topologies so that the links among cameras can be established. 
These scenarios are more realistic but very challenging, as they require the joint efforts of visual-spatio-temporal reasoning. 
Nonetheless, as people usually move at slow speeds and the gaps between camera views are small, their association in the spatio-temporal domain is relatively easy.

VeRi-776~\cite{Liu16a} has been the most widely used benchmark for vehicle ReID, because of the high quality of annotations and the availability of camera geometry. 
However, the dataset does not provide the original videos and calibration information for MTMC tracking purposes. 
Furthermore, the dataset only contains scenes from a city highway, so the variation between viewpoints is rather limited. 
Last but not least, they implicitly make the assumption that MTSC tracking works perfectly. 
As for the other benchmarks~\cite{Liu16b, Yan17}, they are designed for image-level comparison with front and back views only.
Since many vehicles share the same models and different vehicle models can look highly similar, the solution in vehicle ReID should not rely on appearance features only. 
It is important to leverage the spatio-temporal information to address the city-scale problem properly. 
The research community is in urgent need for a benchmark enabling MTMC vehicle tracking analysis. 
\section{\proposed\ benchmark}

In this section, we detail the statistics of the proposed benchmark. 
We also explain how the data were collected and annotated, as well as how we evaluated our baselines. 

\subsection{Dataset overview}

The proposed dataset contains 3.25 hours of videos collected from 40 cameras spanning across 10 intersections in a mid-sized U.S. city. 
The distance between the two furthest simultaneous cameras is 2.5 km, which is the longest among all the existing benchmarks.
The dataset covers a diverse set of location types, including intersections, stretches of roadways, and highways.
With the largest spatial coverage and diverse scenes and traffic conditions, it is the first benchmark that enables city-scale video analytics. 
The benchmark also provides the first public dataset supporting MTMC tracking of vehicles. 

\begin{figure}[t]
\begin{center}
\includegraphics[width=1.0\linewidth]{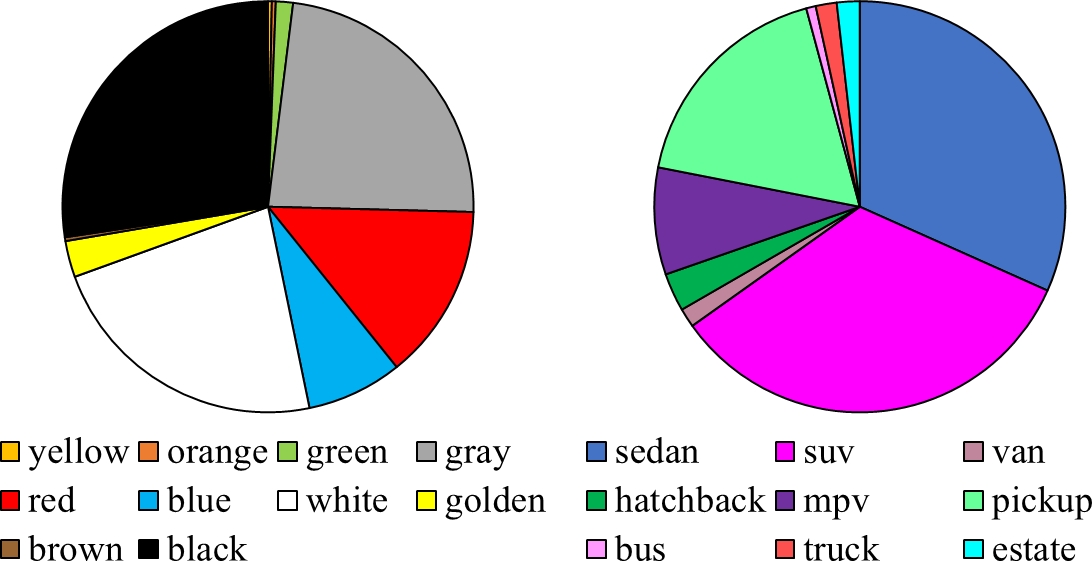}
\end{center}
   \caption{The distribution of vehicle colors and types in terms of vehicle identities in \proposed.}
\label{fig:distribution}
\end{figure}

\begin{figure}[t]
\begin{center}
\includegraphics[width=1.0\linewidth]{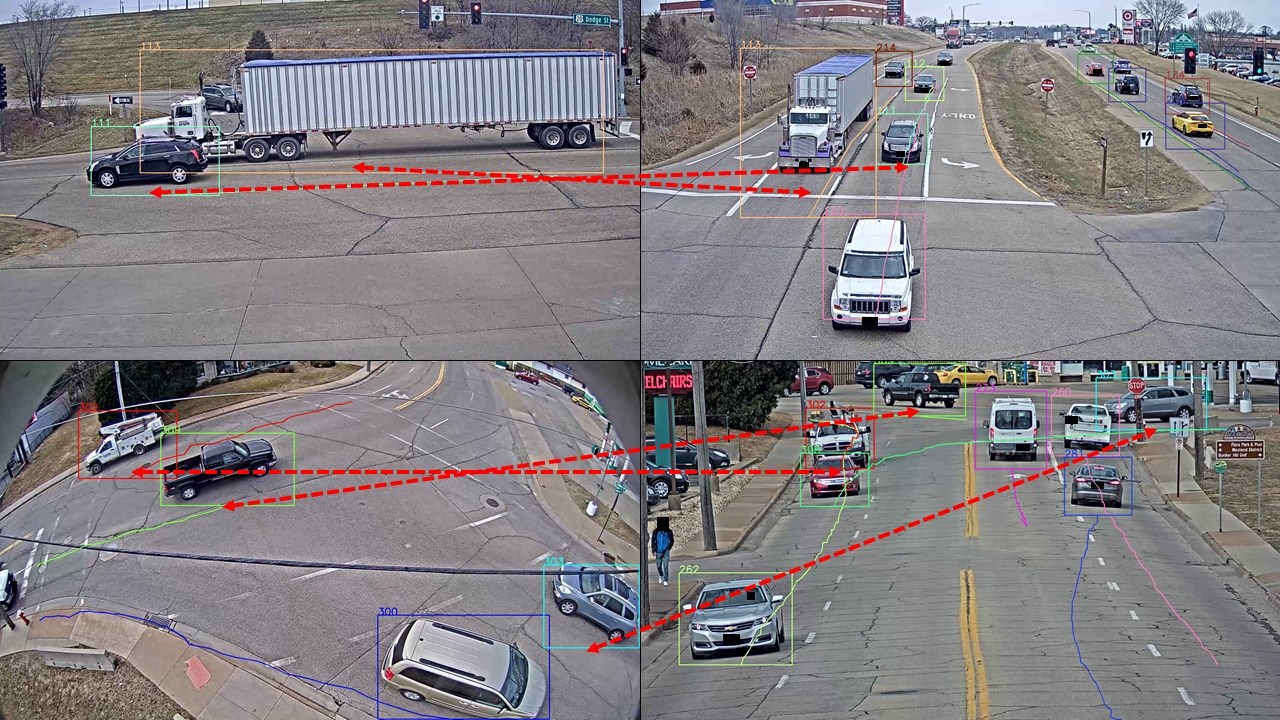}
\end{center}
   \caption{Annotations on \proposed , with red dashed lines indicating associations of object identities across camera views.}
\label{fig:annotation}
\end{figure}

The dataset is divided into 5 scenarios, summarized in Tab.~\ref{tab:scenario}. 
In total, there are 229,680 bounding boxes of 666 vehicle identities annotated, where each passes through at least 2 cameras.
The distribution of vehicle types and colors in \proposed\ is displayed Fig.~\ref{fig:distribution}. 
The resolution of each video is at least 960p and the majority of the videos have a frame rate of 10 FPS. 
Additionally, in each scenario, the offset of starting time for each video is available, which can be used for synchronization. 
For privacy concerns, license plates and human faces detected by DeepStream~\cite{DeepStream} have been redacted and manually refined in all videos.  
\proposed\ also shows other challenges not present in the person-based MTMC tracking benchmarks~\cite{Ristani16, Wu18, Chen15}. Cameras at the same intersection sometimes share overlapping field of views (FOVs) and some cameras use fish-eye lens, leading to strong radial distortion of their captured footage. 
Besides, because of the relatively fast vehicle speed, motion blur may lead to failures in object detection and data association. 
Fig.~\ref{fig:annotation} shows an example of our annotations in the benchmark.
The dataset will be expanded to include more data in diverse conditions in the near future.

\subsection{Data annotation}

To efficiently label tracks of vehicles across multiple cameras, a trajectory-level annotation scheme was employed. 
First, we followed the tracking-by-detection paradigm and generated noisy trajectories in all videos using the state-of-the-art methods in object detection~\cite{Redmon18} and MTSC tracking~\cite{Tang18a}. 
The detection and tracking errors, including misaligned bounding boxes, false negatives, false positives and identity switches, were then manually corrected. 
Finally, we manually associated trajectories across cameras using spatio-temporal cues. 

\begin{figure}[t]
\begin{center}
\includegraphics[width=1.0\linewidth]{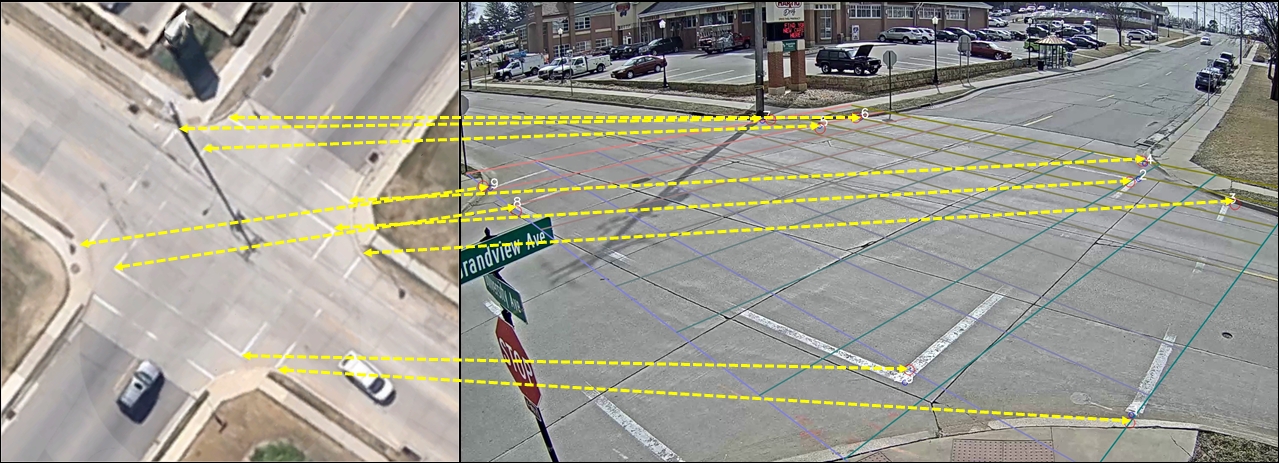}
\end{center}
   \caption{Camera calibration, including manually selecting landmark points in the perspective image (right) and the top-down map view with GPS coordinates (left). The yellow dashed lines indicate the association between landmark points, whereas thin colored solid lines show a ground plane grid projected onto the image using the estimated homography.}
\label{fig:cal}
\end{figure}

The camera geometry of each scenario is available with the dataset. 
We also provide the camera homography matrices between the 2D image plane and the ground plane defined by GPS coordinates based on the flat-earth approximation. 
The demonstration of camera calibration is shown in Fig.~\ref{fig:cal}, which estimates the homography matrix based on the correspondence between a set of 3D points and their 2D pixel locations. 
First, 5 to 14 landmark points were manually selected in a sampled frame image from each video.
Then, the corresponding GPS coordinates in the real world were derived from Google Maps~\cite{GoogleMaps}. 
The objective cost function in this problem is the reprojection error in pixels, where the targeted homography matrix has 8 degrees of freedom. 
This optimization problem can be effectively solved by methods like least median of squares and RANSAC. 
In our benchmark, the converged reprojection error was 11.52 pixels on average, caused by the limited precision of Google Maps. 
When a camera is under radial distortion, it is first manually corrected by straightening curved traffic lane lines before camera calibration. 

\begin{table}
\begin{footnotesize}
\begin{center}
\begin{tabular}{lcccccc}
\toprule
 & Time (min.) & \# cam.\ & \# boxes & \# IDs & Scene type & LOS \\
\midrule
1 & 17.13 & 5 & 20,772 & 95 & highway & A \\
2 & 13.52 & 4 & 20,956 & 145 & highway & B \\
3 & 23.33 & 6 & 6,174 & 18 & residential & A \\
4 & 17.97 & 25 & 17,302 & 71 & residential & A \\
5 & 123.08 & 19 & 164,476 & 337 & residential & B \\
\addlinespace
\multicolumn{2}{l}{total \hspace{1.7ex} 195.03} & 40 & 229,680 & 666 &  &  \\
\bottomrule
\end{tabular}
\end{center}
\caption{The 5 scenarios in the proposed dataset, showing the total time, numbers of cameras (some are shared between scenarios), bounding boxes, and identities, as well as the scene type (highways or residential areas/city streets), and traffic flow (using the North American standard for level of service (LOS)  \cite{FHWA04}). Scenarios 1, 3, and 4 are used for training, whereas 2 and 5 are for testing.}
\label{tab:scenario}
\end{footnotesize}
\end{table} 

\subsection{Subset for image-based ReID} \label{sec:imagereid}

A sampled subset from \proposed , noted as \proposed -ReID, is dedicated for the task of image-based ReID.
\proposed -ReID contains 56,277 bounding boxes in total, where 36,935 of them from 333 object identities form the training set, and the test set consists of 18,290 bounding boxes from the other 333 identities. 
The rest of the 1,052 images are the queries. 
On average, each vehicle has 84.50 image signatures from 4.55 camera views. 

\subsection{Evaluation server}

An online evaluation server is launched with the release of our benchmark at the \href{https://www.aicitychallenge.org/}{2019 AI City Challenge}. 
This allows for continuous evaluation and year-round submission of results against the benchmark. 
A leader board is presented ranking the performances of all submitted results. 
A common evaluation methodology based on the same ground truths ensures fair comparison.
Besides, the state-of-the-art can be conveniently referred to by the research community. 

\subsection{Experimental setup and evaluation metrics} \label{sec:expmetric}

For the evaluation of image-based ReID, the results are represented by a matrix mapping each query to the test images ranked by distance. 
Following~\cite{Zheng15a}, two metrics are used to evaluate the accuracy of algorithms: mean Average Precision (mAP), which measures the mean of all queries' average precision (the area under the Precision-Recall curve), and the rank-$K$ hit rate, denoting the possibility that at least one true positive is ranked within the top $K$ positions. 
In our evaluation server, due to limited storage space, the mAP measured by the top 100 matches for each query is adopted for comparison. 
More details are provided in the supplementary material. 

As for the evaluation of MTMC tracking, we adopted the metrics used by the MOTChallenge~\cite{Bernardin08, Li09} and DukeMTMC~\cite{Ristani16} benchmarks. 
The key measurements include the Multiple Object Tracking Accuracy (MOTA), Multiple Object Tracking Precision (MOTP), ID F1 score (IDF1), mostly tracked targets (MT) and false alarm rate (FAR). 
MOTA computes the accuracy considering three error sources: false positives, false negatives/missed targets and identity switches. 
On the other hand, MOTP takes into account the misalignment between the annotated and the predicted bounding boxes. 
IDF1 measures the ratio of correctly identified detections over the average number of ground-truth and computed detections. 
Compared to MOTA, IDF1 helps resolve the ambiguity among error sources.
MT is the ratio of ground-truth trajectories that are covered by track hypotheses for at least 80\% of their respective life span. 
Finally, FAR measures the average number of false alarms per image frame. 

\section{Evaluated baselines}

This section describes the state-of-the-art baseline systems that we evaluated using the \proposed\ benchmark.

\subsection{Image-based ReID}

For the person ReID problem, the state-of-the-art apply metric learning with different loss functions, such as hard triplet loss (\textbf{Htri})~\cite{Hermans17}, cross entropy loss (\textbf{Xent})~\cite{Szegedy16}, center loss (\textbf{Cent})~\cite{Wen16}, and their combination to train classifiers~\cite{Zhou18a}. 
In our experiments, we compared the performance of various convolutional neural network (CNN) models~\cite{He16, Yu17, Hu18, Xie17, Huang17, Szegedy17, Sandler18}, which are all trained using the same learning rate (3e-4), number of epochs (60), batch size (32), and optimizer (Adam). 
All the trained models fully converge under these hyper-parameter settings.
The generated feature dimension is between 960 and 3,072.

For the vehicle ReID problem, the recent work~\cite{Kumar19} explores the advances in batch-based sampling for triplet embedding that are used for state-of-the-art in person ReID solutions. 
They compared different sampling variants and demonstrated state-of-the-art results on all vehicle ReID benchmarks~\cite{Liu16a, Liu16b, Yan17}, outperforming multi-view-based embedding and most spatio-temporal regularizations (see Tab.~\ref{tab:vehreid}). 
Chosen sampling variants include batch all (\textbf{BA}), batch hard (\textbf{BH}), batch sample (\textbf{BS}) and batch weighted (\textbf{BW}), adopted from~\cite{Hermans17, Ristani18}. 
The implementation uses MobileNetV1~\cite{Howard17} as the backbone neural network architecture, setting the feature vector dimension to 128, the learning rate to 3e-4, and the batch size to $18\times4$.

Another state-of-the-art vehicle ReID method~\cite{Tang18a} is the winner of the vehicle ReID track in the AI City Challenge Workshop at CVPR 2018~\cite{Naphade18}, which is based on fusing visual and semantic features (\textbf{FVS}). 
This method extracts 1,024-dimension CNN features from a GoogLeNet~\cite{Szegedy15} pre-trained on the CompCars benchmark~\cite{Yang15}. 
Without metric learning, the Bhattacharyya norm is used to compute the distance between pairs of feature vectors. 
In our experiments, we also explored the use of the L\textsubscript{2} norm, L\textsubscript{1} norm and L\textsubscript{$\infty$} norm for proximity computations.

\subsection{Single-camera tracking and object detection}

Most state-of-the-art MTSC tracking methods follow the tracking-by-detection paradigm.
In our experiments, we first generate detected bounding boxes using well-known methods such as \textbf{YOLOv3}~\cite{Redmon18}, \textbf{SSD512}~\cite{Liu16c} and \textbf{Faster R-CNN}~\cite{Ren15}. 
For all detectors, we use default models pre-trained on the COCO benchmark~\cite{Lin14}, where the classes of interest include car, truck and bus. 
We also use the same threshold for detection scores across all methods (0.2). 

Offline methods in MTSC tracking usually lead to better performance, as all the aggregated tracklets can be used for data association. Online approaches often leverage robust appearance features to compensate for not having information about the future. 
We experimented with both types of methods in \proposed, which are introduced as follows.
\textbf{DeepSORT}~\cite{Wojke17} is an online method that combines deep learning features with Kalman-filter-based tracking and the Hungarian algorithm for data association, achieving remarkable performance on the MOTChallenge MOT16 benchmark~\cite{Milan16}. 
\textbf{TC}~\cite{Tang18a} is an offline method that won the traffic flow analysis task in the AI City Challenge Workshop at CVPR 2018~\cite{Naphade18} by applying tracklet clustering through optimizing a weighted combination of cost functions, including smoothness loss, velocity change loss, time interval loss and appearance change loss.
Finally, \textbf{MOANA}~\cite{Tang19b, Tang18b} is another online method that achieves state-of-the-art performance on the MOTChallenge 2015 3D benchmark~\cite{Leal-Taixé15}, employing similar schemes for spatio-temporal data association, but using an adaptive appearance model to resolve occlusion and grouping of objects.

\subsection{Spatial-temporal analysis}

The intuition behind spatio-temporal association is that the moving patterns of vehicles are predictable, because they usually follow traffic lanes, and the speed changes smoothly.  
Liu \etal~\cite{Liu17} propose a progressive and multimodal vehicle ReID framework (\textbf{PROVID}), in which a spatio-temporal-based re-ranking scheme is employed. 
The spatio-temporal similarity is measured by computing the ratios of time difference and physical distance across cameras.

More sophisticated algorithms apply probabilistic models to learn the transition between pairs of cameras. 
For example, a method based on two-way Gaussian mixture model features (\textbf{2WGMMF})~\cite{Lee18} achieves state-of-the-art accuracy on the NLPR\_MCT benchmark~\cite{Chen15} by learning the transition time between camera views using Gaussian distributions. 
In \textbf{FVS}~\cite{Tang18a}, however, since no training data is provided, the temporal distribution is pre-defined based on the estimated distance between cameras.
Both methods require manual selection of entry/exit zones in camera views, but 2WGMMF can learn the camera link model online. 
\section{Experimental evaluation results}

In this section we analyze the performance of various state-of-the-art methods on our \proposed\ benchmark and compare our benchmark to existing ones.

\begin{table}
\begin{footnotesize}
\begin{center}
\begin{tabular}{ccccc}
\toprule
Norm & mAP & Rank-1 & Rank-5 & Rank-10\\
\midrule
Bhattacharyya & \textbf{6.3\%} & \textbf{20.8\%} & 24.5\% & \textbf{27.9\%} \\
L$_{2}$ & 5.9\% & 20.4\% & \textbf{24.9\%} & \textbf{27.9\%} \\
L$_{1}$ & 6.2\% & 20.3\% & 24.8\% & 27.8\% \\
L$_{\infty}$ & 3.2\% & 17.0\% & 23.6\% & 27.6\% \\
\bottomrule
\end{tabular}
\end{center}
\caption{Performance of CNN features extracted from a leading vehicle ReID method, FVS~\cite{Tang18a}, compared using various metrics, on our \proposed -ReID benchmark.}
\label{tab:reidfvs}
\end{footnotesize}
\end{table}

\begin{figure}[t]
\begin{center}
\includegraphics[width=1.0\linewidth]{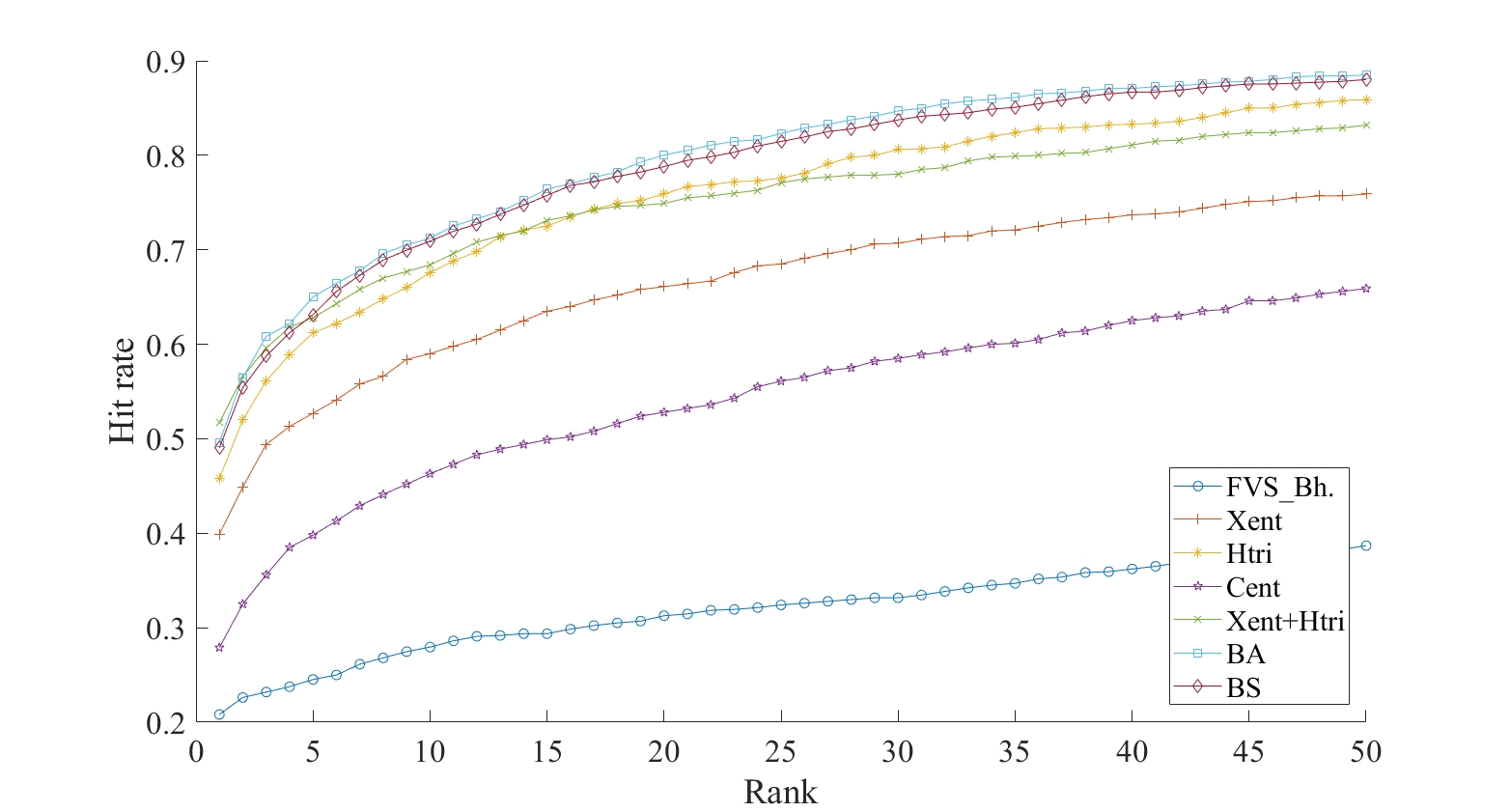}
\end{center}
   \caption{CMCs of image-based ReID methods on \proposed -ReID. DenseNet121~\cite{Huang17} is used for all the state-of-the-art person ReID schemes in Tab.~\ref{tab:reidmetlrn}.}
\label{fig:cmc}
\end{figure}

\subsection{Image-based ReID}

First, we evaluate the performance of state-of-the-art ReID methods on \proposed -ReID, which is the subset of our benchmark for image-based ReID mentioned in Section~\ref{sec:imagereid}.  
Our goal is to determine whether \proposed -ReID is challenging for existing methods.

\textbf{Non-metric learning method.}    
The deep features output by a CNN can be directly compared using standard distance metrics.
Tab.~\ref{tab:reidfvs} shows the results of the FVS method~\cite{Tang18a} using various distance metrics. 
Overall, the performance of non-metric learning is poor. Furthermore, the model is pre-trained on a dataset for fine-grained vehicle classification~\cite{Yang15}, which would hurt some performance gains versus pre-training on vehicle ReID dataset. 



\begin{figure}[t]
\begin{center}
\includegraphics[width=0.85\linewidth]{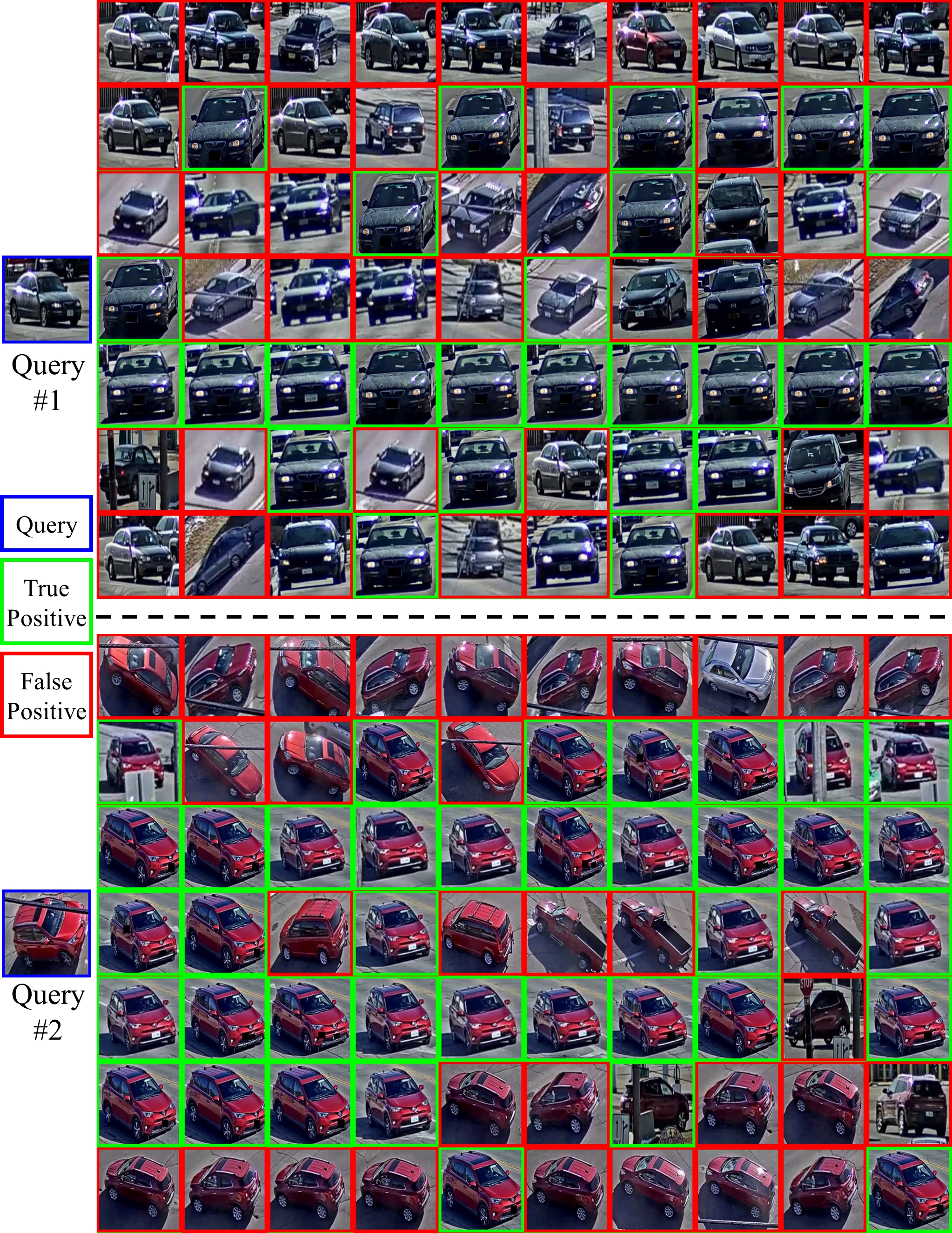}
\end{center}
   \caption{Qualitative performance of image-based ReID methods for two example queries from \proposed -ReID. The rows of each query show, from top to bottom, the results of FVS (Bhattacharyya norm), Xent, Htri, Cent, Xent+Htri, BA and BS. Each row shows the top 10 matches found by that method. DenseNet121~\cite{Huang17} is used for all the state-of-the-art person ReID schemes in Tab.~\ref{tab:reidmetlrn}.}
\label{fig:visreid}
\end{figure}

\begin{table*}
\begin{footnotesize}
\begin{center}
\begin{tabular}{lcccccccc}
\toprule
Loss & ResNet50 & ResNet50M & ResNeXt101 & SEResNet50 & SEResNeXt50 & DenseNet121 & InceptionResNetV2 & MobileNetV2 \\
& \cite{He16} & \cite{Yu17} & \cite{Xie17} & \cite{Hu18} & \cite{Hu18} & \cite{Huang17} & \cite{Szegedy17} & \cite{Sandler18} \\
\midrule
Xent~\cite{Szegedy16} & 25.5 (41.3) & 25.3 (42.1) & 26.6 (42.4) & 23.8 (40.4) & \textbf{26.8} (\textbf{45.2}) & 23.2 (39.9) & 20.8 (35.5) & \textbf{14.7} (\textbf{26.0}) \\
Htri~\cite{Hermans17} & 28.7 (42.9) & 27.9 (40.1) & 30.0 (41.3) & 26.3 (38.7) & 28.2 (40.4) & \textbf{30.5} (\textbf{45.8}) & 23.7 (37.2) & 0.4 (0.3) \\
Cent~\cite{Wen16} & 7.6 (18.2) & 7.9 (21.5) & 8.1 (19.3) & 10.0 (25.9) & 10.2 (25.6) & \textbf{10.7} (\textbf{27.9}) & 6.0 (15.2) & 7.9 (18.4) \\
Xent+Htri & \textbf{29.4} (\textbf{45.9}) & \textbf{29.4} (\textbf{49.7}) & \cellcolor{lightgray}\textbf{32.0} (\textbf{48.8}) & \textbf{30.0} (\textbf{47.2}) & \textbf{30.8} (\textbf{49.1}) & \cellcolor{lightgray}\textbf{31.0} (\textbf{51.7}) & \textbf{25.6} (\textbf{42.2}) & 11.2 (16.3) \\
Xent+Cent & 23.1 (37.5) & 26.5 (47.3) & 24.9 (40.9) & 26.2 (43.7) & \textbf{28.4} (47.5) & 27.8 (\textbf{48.1}) & 23.5 (39.5) & 12.3 (24.0) \\
\bottomrule
\end{tabular}
\end{center}
\caption{State-of-the-art metric learning methods for person ReID on \proposed -ReID, showing mAP and rank-1 (in parentheses), as percentages. All networks were pre-trained on ImageNet~\cite{Deng09}.  The best architecture and loss function are highlighted for each row/column, respectively, with the shaded cells indicating the overall best for both mAP and rank-1. }
\label{tab:reidmetlrn}
\end{footnotesize}
\end{table*}

\textbf{Metric learning methods in person ReID.}  
Tab.~\ref{tab:reidmetlrn} shows results of state-of-the-art metric learning methods for person ReID on the \proposed -ReID dataset,
using different loss functions and network architectures.
The performance is much improved compared to the non-metric learning method in Tab.~\ref{tab:reidfvs}. 
In particular, hard triplet loss is the most robust. 
A combination of hard triplet loss and cross-entropy loss yields the best results.
As for CNN architectures, DenseNet121~\cite{Huang17} achieves the highest accuracy in most cases, as it benefits from improved flow of information and gradients throughout the network. 

\textbf{Person ReID methods on other benchmarks.}
Despite the above efforts to explore network architectures and combine metric learning losses, the top mAP on our \proposed -ReID benchmark is still lower than 35\%. 
In comparison, Tab.~\ref{tab:personreid}~\cite{Zhou18a, SotaMarket1501, SotaDukeMTMC, SotaMSMT17} shows the performance of the same methods on other public benchmarks, using the same implementations and hyperparameters.
In general, performance is significantly better, thus verifying that \proposed -ReID is indeed more challenging. 

\begin{table}
\begin{footnotesize}
\begin{center}
\begin{tabular}{ccccc}
\toprule
Method & Market1501 & DukeMTMC-reID & MSMT17 \\
& \cite{Zheng15a} & \cite{Ristani16, Zheng17a} & \cite{Wei18a} \\
\midrule
HA-CNN~\cite{Li18} & 75.6 (90.9) & 63.2 (80.1) & 37.2 (64.7) \\
MLFN~\cite{Chang18} & 74.3 (90.1) & 63.2 (81.1) & 37.2 (66.4) \\
GLAD~\cite{Wei17} & - & - & 34.0 (61.4) \\
\addlinespace
Res50+Xent & 75.3 (90.8) & \textbf{64.0} (81.0) & 38.4 (69.6) \\
Res50M+Xent & \textbf{76.0} (90.2) & \textbf{64.0} (\textbf{81.6}) & 38.0 (69.0) \\
SERes50+Xent & 75.9 (\textbf{91.9}) & 63.7 (81.5) & \textbf{39.8} (\textbf{71.1}) \\
Dense121+Xent & 68.0 (87.8) & 58.8 (79.7) & 35.0 (67.6) \\
\bottomrule
\end{tabular}
\end{center}
\caption{State-of-the-art metric learning methods for person ReID on other public benchmarks, showing mAP and rank-1 (in parentheses), as percentages. The bottom rows (from~\cite{Zhou18a}) show that the methods from Tab.~\ref{tab:reidmetlrn} are competitive against the state-of-the-art.}
\label{tab:personreid}
\end{footnotesize}
\end{table}

\begin{table}
\begin{footnotesize}
\begin{center}
\begin{tabular}{lcccc}
\toprule
Method & mAP & Rank-1 & Rank-5 & Rank-10\\
\midrule
MoV1+BA~\cite{Kumar19} & 31.3\% & 49.6\% & 65.0\% & 71.2\% \\
MoV1+BH~\cite{Kumar19} & \textbf{32.0\%} & 48.4\% & \textbf{65.2\%} & \textbf{71.4\%} \\
MoV1+BS~\cite{Kumar19} & 31.3\% & 49.0\% & 63.1\% & 70.9\% \\
MoV1+BW~\cite{Kumar19} & 30.8\% & \textbf{50.1\%} & 64.9\% & \textbf{71.4\%} \\
\bottomrule
\end{tabular}
\end{center}
\caption{The state-of-the-art metric learning method for vehicle ReID, with different sampling variants, on \proposed -ReID.}
\label{tab:reidemb}
\end{footnotesize}
\end{table}

\begin{table}
\begin{footnotesize}
\begin{center}
\begin{tabular}{lcccc}
\toprule
Method & VeRi-776 & VehicleID & PKU-VD1 & PKU-VD2 \\
& \cite{Liu16a} & \cite{Liu16b} & \cite{Yan17} & \cite{Yan17} \\ 
\midrule
GSTE~\cite{Bai18} & 59.5 (96.2) & 72.4 (74.0) & - & - \\
VAMI~\cite{Zhou18b} & 50.1 (77.0) & - (47.3) & - & - \\
OIFE~\cite{Wang17} & 48.0 (89.4) & - (67.0) & - & - \\
CCL~\cite{Liu16b} & - & 45.5 (38.2) & - & - \\
MGR~\cite{Yan17} & - & - & 51.1 (-) & 55.3 (-) \\
\addlinespace
MoV1+BA~\cite{Kumar19} & 66.9 (90.1) & 76.0 (66.7) & - & - \\
MoV1+BH~\cite{Kumar19} & 65.1 (87.3) & 76.9 (67.6) & - & - \\
MoV1+BS~\cite{Kumar19} & \textbf{67.6 (90.2)} & \textbf{78.2 (69.3)} & \textbf{58.3 (58.3)} & \textbf{62.4 (69.4)} \\
MoV1+BW~\cite{Kumar19} & 67.0 (90.0) & 78.1 (69.4) & - & - \\
\bottomrule
\end{tabular}
\end{center}
\caption{State-of-the-art metric learning methods for vehicle ReID on other public benchmarks, showing mAP and rank-1 (in parentheses), as percentages. Performance is evaluated on the largest test sets for VehicleID, PKU-VD1 and PKU-VD2. The bottom rows show the methods in our comparison (from Tab.~\ref{tab:reidemb}).}
\label{tab:vehreid}
\end{footnotesize}
\end{table}

\begin{table}
\begin{footnotesize}
\begin{center}
\begin{tabular}{lcccccc}
\toprule
Method & IDF1 & Recall & FAR & MT & MOTA & MOTP\\
\midrule
DS+YOLO & 78.9\% & 67.6\% & 8.6 & 778 & 67.4\% & 65.8\% \\
DS+SSD & 79.5\% & 69.2\% & 8.3 & 756 & 68.9\% & 65.5\% \\
DS+FRCNN & 78.9\% & 66.9\% & 15.3 & 761 & 66.7\% & 65.5\% \\
\addlinespace
TC+YOLO & 79.1\% & 68.1\% & 8.5 & 871 & 68.0\% & \textbf{66.0\%} \\
TC+SSD & \textbf{79.7\%} & \textbf{70.4\%} & 7.4 & 895 & \textbf{70.3\%} & 65.6\% \\
TC+FRCNN & 78.7\% & 68.5\% & 12.0 & 957 & 68.4\% & 65.9\% \\
\addlinespace
MO+YOLO & 77.8\% & 69.0\% & 8.5 & 965 & 68.6\% & \textbf{66.0\%} \\
MO+SSD & 72.8\% & 68.0\% & \textbf{6.3} & 980 & 67.0\% & 65.9\% \\
MO+FRCNN & 75.6\% & 69.5\% & 10.8 & \textbf{1094} & 68.6\% & \textbf{66.0\%} \\
\bottomrule
\end{tabular}
\end{center}
\caption{State-of-the-art methods for MTSC tracking and object detection on \proposed. 
The metrics are explained in Section~\ref{sec:expmetric}.
}
\label{tab:mtsc}
\end{footnotesize}
\end{table}
\begin{table*}
\begin{footnotesize}
\begin{center}
\begin{tabular}{lccccccccc}
\toprule
\multirow{2}{1in}{Spatio-temporal association} & \multirow{2}{0.5in}{MTSC tracking} & \multicolumn{6}{c}{Image-based ReID} & \\
\cmidrule{3-9}
 &  & FVS\_Bh. & Xent & Htri & Cent & Xent+Htri & BA & BS\\
\midrule
\multirow{3}{*}{\rotatebox{0}{PROVID~\cite{Liu17}}} &
DeepSORT~\cite{Wojke17} & 21.5\% & 31.3\% & 35.3\% & 27.6\% & 34.5\% & \textbf{35.6\%} & 33.6\% \\
& TC~\cite{Tang18a} & 22.1\% & 35.2\% & 39.4\% & 32.7\% & 39.9\% & \textbf{40.6\%}  & 39.0\% \\
& MOANA~\cite{Tang19b} & 21.7\% & 29.1\% & 33.0\% & 26.1\% & 31.9\% & \textbf{34.4\%} & 31.8\% \\
\addlinespace
\multirow{3}{*}{\rotatebox{0}{2WGMMF~\cite{Lee18}}} & 
DeepSORT~\cite{Wojke17} & 25.0\% & 35.3\% & 38.4\% & 31.2\% & 37.5\% & \textbf{40.3\%} & 39.8\% \\
& TC~\cite{Tang18a} & \textbf{27.6\%} & 39.5\% & 41.7\% & 34.7\% & \textbf{43.3\%} & 44.1\% & \textbf{45.1\%} \\
& MOANA~\cite{Tang19b} & 20.2\% & 32.2\% & 35.9\% & 28.2\% & 36.5\% & \textbf{38.1\%} & 37.7\% \\
\addlinespace
\multirow{3}{*}{\rotatebox{0}{FVS~\cite{Tang18a}}} &
DeepSORT~\cite{Wojke17} & 24.9\% & 36.4\% & 40.0\% & 30.8\% & 39.0\% & 41.3\% & \textbf{41.4\%} \\
& TC~\cite{Tang18a} & \textbf{27.6\%} & \textbf{40.5\%} & \textbf{42.7\%} & \textbf{36.6\%} & 42.4\% & \cellcolor{lightgray}\textbf{46.3\%} & \textbf{46.0\%} \\
& MOANA~\cite{Tang19b} & 21.2\% & 32.7\% & 36.4\% & 29.2\% & 37.5\% & \textbf{39.5\%} & 36.9\% \\
\bottomrule
\end{tabular}
\end{center}
\caption{MTMC tracking with different combinations of spatio-temporal association, MTSC tracking (supported by SSD512~\cite{Liu16c}), and image-based ReID methods on \proposed . Each cell shows the ID F1 score. The best performance is highlighted for each row/column, with the shaded cells indicating the overall best. DenseNet121~\cite{Huang17} is used for the comparison of Xent, Htri, Cent and Xent+Htri. }
\label{tab:mtmc}
\end{footnotesize}
\end{table*}

\textbf{Metric learning methods in vehicle ReID.}
Tab.~\ref{tab:reidemb} displays the results of the state-of-the-art for vehicle ReID~\cite{Kumar19} on the proposed dataset. 
For this experiment we compare sampling variants (BA, BH, BS, and BW) using an implementation based on MobileNetV1~\cite{Howard17}, as described earlier. 

Results in terms of rank-1 hit rates are only slightly worse than those from the combination of hard triplet loss and cross-entropy loss in person ReID (see Tab.~\ref{tab:reidmetlrn}).  
This reduction in precision is likely due to the relatively simple network architecture (MobileNetV1~\cite{Howard17}) and a computationally efficient embedding into 128 dimensions.
Tab.~\ref{tab:reidemb} demonstrates yet again the challenges of \proposed -ReID.


\textbf{Vehicle ReID methods on other benchmarks.}
To verify that our method is indeed competitive, Tab.~\ref{tab:vehreid}~\cite{Kumar19} shows the performance of several state-of-the-art vehicle ReID approaches on public benchmarks. 
%
%

These results are also summarized in the cumulative match curve (CMC) plots in Fig.~\ref{fig:cmc}. 
Qualitative visualization of performance is shown in Fig.~\ref{fig:visreid}.
We observe that most failures are caused by viewpoint variations, which is a key problem that should be addressed by future methods. 



\subsection{MTSC tracking and object detection}

Reliable cross-camera tracking is built upon accurate tracking within each camera (MTSC). 
Tab.~\ref{tab:mtsc} shows results of state-of-the-art methods for MTSC tracking~\cite{Wojke17, Tang19b, Tang18a} combined with leading object detection algorithms~\cite{Redmon18, Liu16c, Ren15} on \proposed. 
Note that false positives are not taken into account in MTSC tracking evaluation, because only vehicles that travel across more than one camera are annotated. 
With regards to object detectors, SSD512~\cite{Liu16c} performs the best, whereas YOLOv3~\cite{Redmon18} and Faster R-CNN~\cite{Ren15} show similar performance.
As for MTSC trackers, TC~\cite{Tang18a}, the only offline method, performs better according to most of the evaluation metrics. 
DeepSORT~\cite{Wojke17} and MOANA~\cite{Tang19b} share similar performance in MOTA, but the ID F1 scores of DeepSORT are much higher.
Nonetheless, MOANA is capable of tracking most trajectories successfully.

\subsection{MTMC tracking}

MTMC tracking is a joint process of visual-spatio-temporal reasoning. 
For these experiments, we first apply MTSC tracking, then sample a number of signatures from each trajectory in order to extract and compare appearance features. 
The number of sampled instances from each vehicle is empirically chosen as 3.

Tab~\ref{tab:mtmc} shows the results of various methods for spatio-temporal association, MTSC tracking, and image-based ReID on \proposed.
Note that PROVID~\cite{Liu17} compares visual features first, then uses spatio-temporal information for re-ranking; whereas 2WGMMF~\cite{Lee18} and FVS~\cite{Tang18a} first model the spatio-temporal transition based on online learning or manual measurements, and then perform image-based ReID only on the confident pairs. 
Note also that, since only trajectories spanning multiple cameras are included in the evaluation, different from MTSC tracking, false positives are considered in the calculation of MTMC tracking accuracy. 

Overall the most reliable spatio-temporal association method is FVS, which exploits a manually specified probabilistic model of transition time. 
In comparison, 2WGMMF achieves performance comparable with FVS in most cases, due to the online learned transition time distribution applied to those cameras that are shared between the training and test sets.
Without probabilistic modeling, PROVID yields inferior results. 
We can also conclude from Tab~\ref{tab:mtmc} that the choice of image-based ReID and MTSC tracking methods has a significant impact on overall performance, as those methods achieving superior performance in their sub-tasks also contribute to higher MTMC tracking accuracy. 

\section{Conclusion}

We proposed a city-scale benchmark, \proposed, which enables both video-based MTMC tracking and image-based ReID tasks. 
Our major contribution is three-fold. 
First, \proposed\ is the first attempt towards city-scale applications in traffic understanding.
It has the largest scale among all the existing ReID datasets in terms of spatial coverage and the number of cameras/intersections involved. 
Moreover, a wide range of scenes and traffic flow conditions are included. 
Second, \proposed\ is also the first benchmark to support vehicle-based MTMC tracking, by providing annotations for the original videos, the camera geometry, and calibration information. 
The provided spatio-temporal information can be leveraged to resolve ambiguity in image-based ReID. 
Third, we conducted extensive experiments evaluating the performance of state-of-the-art approaches on our benchmark, comparing and analyzing various visual-spatio-temporal association schemes. 
We show that our scenarios are challenging and reflect realistic situations that deployed systems will need to operate in.
Finally, \proposed\ may also open the way for new research problems such as vehicle pose estimation, viewpoint generation, \textit{etc.}


{\small
\bibliographystyle{ieee_fullname}
\bibliography{main}
}

\clearpage
\section{Supplementary}

This supplementary material includes additional details regarding the definitions of the evaluation metrics for MTMC tracking as well as MTSC tracking, which are partially explained in Section~\ref{sec:expmetric}. 
The measurements are adopted from the MOTChallenge~\cite{Bernardin08, Li09} and DukeMTMC~\cite{Ristani16} benchmarks. 
Besides, the performance of our baseline image-based ReID methods in terms of mAP measured by the top 100 matches for each query is presented, which is the metric used in our evaluation server.

\subsection{Metrics in CLEAR MOT}
The CLEAR MOT~\cite{Bernardin08} metrics are used in the MOTChallenge benchmark for evaluating multiple object tracking performance. 
The distance measure, \textit{i.e.}, how close a tracker hypothesis is to the actual target, is determined by the intersection over union between estimated bounding boxes and the ground truths. 
The similarity threshold for true positives is empirically set to 50\%.

The Multiple Object Tracking Accuracy (MOTA) combines three sources of errors to evaluate a tracker's performance.
\begin{equation}
\text{MOTA} = 1-\frac{\sum_{t} \left( \text{FN}_{t} + \text{FP}_{t} + \text{IDSW}_{t}\right)}{\sum_{t} \text{GT}_{t}},
\end{equation}
where $t$ is the frame index.
FN, FP, IDSW and GT respectively denote the numbers of false negatives, false positives, identity switches and ground truths. 
The range of MOTA in percentage is $\left( -\infty, 100 \right]$, which becomes negative when the number of errors exceeds the ground truth objects.

Multiple Object Tracking Precision (MOTP) is used to measure misalignment between annotated and predicted object locations, defined as
\begin{equation}
\text{MOTP} = 1-\frac{\sum_{t,i} d_{t,i}}{\sum_{t} c_{t}},
\end{equation}
in which $c_t$ denotes the number of matches and $d_{t,i}$ is the bounding box overlap between target $i$ and the ground truth at frame index $t$. 
According to the analysis in~\cite{Leal-Taixé15}, MOTP shows a remarkably low variation across different methods compared with MOTA.
Therefore, MOTA is considered as a more reliable evaluation for tracking performance. 

Besides MOTA and MOTP, there are other metrics for evaluating the tracking quality. 
MT measures the number of mostly tracked targets that are successfully tracked by at least 80\% of their life span.
On the other hand, ML calculates the number of mostly lost targets that are only recovered for less than 20\% of their total lengths.
All the other targets are classified as partially tracked (PT). 
Furthermore, FAR measures the average number of false alarms, \textit{i.e.}, FN, FP and IDSW, per frame. 

\begin{table}
\begin{footnotesize}
\begin{center}
\begin{tabular}{cc}
\toprule
Norm & Rank-100 mAP\\
\midrule
Bhattacharyya & \textbf{5.1\%} \\
L$_{2}$ & 5.0\% \\
L$_{1}$ & 4.8\% \\
L$_{\infty}$ & 2.5\% \\
\bottomrule
\end{tabular}
\end{center}
\caption{Performance of non-metric learning methods using CNN features extracted from FVS~\cite{Tang18a} on our \proposed -ReID benchmark, showing rank-100 mAP, corresponding to the experimental results of Tab.~\ref{tab:reidfvs}. }
\label{tab:reidfvs_rankkmap}
\end{footnotesize}
\end{table}

\subsection{Metrics in DukeMTMC}
There are three evaluation metrics introduced by the DukeMTMC benchmark, namely identification precision (IDP), identification recall (IDR), and the F1 score IDF1. 
They are defined based on the counts of false negative identities (IDFN), false positive identities (IDFP) and true positive identities (IDTP), which are defined as follows,
\begin{equation}
\text{IDFN} = \sum_{\tau \in \text{AT}} \sum_{t \in \mathcal{T}_{\tau}} m \left( \tau, \gamma_{m} \left( \tau \right), t \right),
\end{equation}
\begin{equation}
\text{IDFP} = \sum_{\gamma \in \text{AC}} \sum_{t \in \mathcal{T}_{\gamma}} m \left( \tau_{m} \left( \gamma \right), \gamma, t \right),
\end{equation}
\begin{equation}
\text{IDTP} = \sum_{\tau \in \text{AT}} \| \tau \| - \text{IDFN} = \sum_{\gamma \in \text{AC}} \| \gamma \| - \text{IDFP},
\end{equation}
where $\tau$ and $\gamma$ respectively denotes the true and computed trajectories, AT and AC are all true and computed identities, and $\mathcal{T}$ represents the set of frame indices $t$ over which the corresponding trajectory extends. 
$\| \cdot \|$ returns the number of detections in a given trajectory.
The expression $m \left( \tau, \gamma, t \right)$ calculates the number of missed detections between $\tau$ and $\gamma$ along time.
We use $\gamma_{m} \left( \tau \right)$ and $\tau_{m} \left( \gamma \right)$ to denote the bipartite match from $\tau$ to $\gamma$ and \textit{vice versa}, respectively. 
Identification precision (recall) is defined as the ratio of computed (true) detections that are correctly identified. 
\begin{equation}
\text{IDP} = \frac{\text{IDTP}}{\text{IDTP} + \text{IDFP}},
\end{equation}
\begin{equation}
\text{IDR} = \frac{\text{IDTP}}{\text{IDTP} + \text{IDFN}}.
\end{equation}
IDF1 is the fraction of correctly identified detections over the average number of true and computed detections. 
\begin{equation}
\text{IDF1} = \frac{2 \cdot \text{IDF1}}{2 \cdot \text{IDF1} + \text{IDFP} + \text{IDFN}}.
\end{equation}

Compared to the metrics in CLEAR MOT, the truth-to-result mapping in IDF1 computation is not frame-by-frame but identity-by-identity for the entire sequence, and the errors of any type are penalized based on binary mismatch. 
Therefore, IDF1 can handle overlapping and disjoint fields of view for the evaluation of MTMC tracking performance, which is a property absent in all previous measures. 

\begin{table*}
\begin{footnotesize}
\begin{center}
\begin{tabular}{lcccccccc}
\toprule
Loss & ResNet50 & ResNet50M & ResNeXt101 & SEResNet50 & SEResNeXt50 & DenseNet121 & InceptionResNetV2 & MobileNetV2 \\
& \cite{He16} & \cite{Yu17} & \cite{Xie17} & \cite{Hu18} & \cite{Hu18} & \cite{Huang17} & \cite{Szegedy17} & \cite{Sandler18} \\
\midrule
Xent~\cite{Szegedy16} & 20.3\% & 20.4\% & \textbf{21.6\%} & 18.6\% & 21.5\% & 18.6\% & 16.2\% & \textbf{10.4\%} \\
Htri~\cite{Hermans17} & 22.1\% & 21.3\% & 23.3\% & 19.8\% & 21.7\% & \textbf{24.0\%} & 17.8\% & 0.0\% \\
Cent~\cite{Wen16} & 5.6\% & 6.1\% & 6.2\% & 8.3\% & 8.4\% & \textbf{9.5\%} & 4.9\% & 5.2\% \\
Xent+Htri & \textbf{23.7\%} & \textbf{24.2\%} & \cellcolor{lightgray}\textbf{26.3\%} & \textbf{24.3\%} & \textbf{25.1\%} & \textbf{26.0\%} & \textbf{20.5\%} & 6.5\% \\
Xent+Cent & 17.8\% & 21.7\% & 19.5\% & 20.9\% & 23.2\% & \textbf{23.3\%} & 18.8\% & 8.3\% \\
\bottomrule
\end{tabular}
\end{center}
\caption{Performance of state-of-the-art metric learning methods for person ReID on \proposed -ReID, showing rank-100 mAP, corresponding to the experimental results of Tab.~\ref{tab:reidmetlrn}.  The best architecture and loss function are highlighted for each row/column, respectively, with the shaded cells indicating the overall best. }
\label{tab:reidmetlrn_rankkmap}
\end{footnotesize}
\end{table*}

\begin{table}
\begin{footnotesize}
\begin{center}
\begin{tabular}{lc}
\toprule
Method & Rank-100 mAP\\
\midrule
MobileNetV1+BA~\cite{Kumar19} & 25.6\% \\
MobileNetV1+BH~\cite{Kumar19} & \textbf{26.5\%} \\
MobileNetV1+BS~\cite{Kumar19} & 25.6\% \\
MobileNetV1+BW~\cite{Kumar19} & 25.4\% \\
\bottomrule
\end{tabular}
\end{center}
\caption{Performance of the state-of-the-art metric learning method for vehicle ReID, with different sampling variants, on \proposed -ReID, corresponding to the experimental results of Tab.~\ref{tab:reidemb}.   Rank-100 mAP is shown. }
\label{tab:reidemb_rankkmap}
\end{footnotesize}
\end{table}

\subsection{Rank-$K$ mAP for evaluating image-based ReID}
As mentioned in Section~\ref{sec:expmetric}, to measure the total mAP of each submission, a distance matrix of dimension $Q \times T$ is required, where $Q$ and $T$ are the numbers of queries and test images, respectively. 
For an evaluation server with many users and each of them is allowed to submit multiple times, such large file size may lead to system instability. 
Thus, we create a new evaluation metric for image-based ReID, named rank-$K$ mAP, that measures the mean of average precision for each query considering only the top $K$ matches, so that the required dimension of each submission can be reduced to $Q \times K$.
Note that $K$ usually needs to be larger than the maximum length of ground-truth trajectories, which is chosen to be 100 for our evaluation. 

Because rank-100 mAP is adopted in our evaluation server, we present here the additional experimental results in Tab.~\ref{tab:reidfvs_rankkmap}, Tab.~\ref{tab:reidmetlrn_rankkmap} and Tab.~\ref{tab:reidemb_rankkmap}, which correspond to Tab.~\ref{tab:reidfvs}, Tab.~\ref{tab:reidmetlrn} and Tab.~\ref{tab:reidemb}, respectively.

\end{document}